# Attention Is Not Relevance:
## Challenging the Causality of Attention in Autonomous Driving

**Luca Lusvarghi, Networked Systems Laboratory, Universidad Miguel Hernández**
**Javier Gozalvez, Networked Systems Laboratory, Universidad Miguel Hernández**

*Attention has become one of the most widely adopted solutions for estimating the relevance of surrounding objects and interpreting an autonomous vehicle's driving decisions. However, its ability to capture the causal impact of individual input elements on an AI model's output has long been discussed in the literature, particularly in the Natural Language Processing (NLP) domain. This work brings this discussion into the autonomous driving domain by analyzing the accuracy of attention in estimating the impact of an input object on an autonomous vehicle's driving, i.e., its relevance. To this end, we compare different attention-based relevance estimation methods against a causal reference of object relevance obtained through counterfactual analysis. We perform our analysis on more than 80,000 synthetic driving scenes using an autonomous vehicle based on PlanT 2.0, a state-of-the-art Transformer-based model for object-level planning. Results show that attention provides a correlational rather than causal indication of object relevance. Attention-based relevance estimation methods do not consistently identify all relevant objects and tend to incorrectly classify as relevant objects with little or no impact on the vehicle's driving. In particular, results show that attention can effectively identify relevant objects in simple single-lane, straight-road scenarios, but its accuracy degrades in more complex settings, such as multi-lane roads and intersections. These findings highlight the limitations of attention as a proxy for object relevance, calling for novel methods able to accurately capture the causal impact of individual objects on the vehicle's driving and distinguish it from correlational associations.*

### Introduction

Autonomous vehicles operate in complex environments populated by heterogeneous and dynamically interacting objects, including other vehicles, motorcycles, pedestrians, and cyclists. While an autonomous vehicle may detect and track a large number of surrounding objects through its onboard sensors, in practice, only a subset of the detected objects is actually relevant to its driving. In [1] and [2], object relevance is defined from a safety-oriented perspective: an object is considered relevant if its missed detection can have safety-critical implications for the ego-vehicle. Similarly, the authors of [3] link object relevance to collision avoidance and consider an object as relevant if its missed detection can lead to a collision. In other works, object relevance is associated with related concepts such as object importance and object risk. For example, [4] uses object importance to refer to objects that the ego-vehicle should consider to make safe driving decisions, while [5] uses the term more broadly to refer to the objects that are relevant to a human driver's decisions. Object risk is instead used in [6], where the authors define risk objects as those that cause a human driver to stop or change lane.

We define the relevance of an object as the impact that the object has on the autonomous vehicle's driving. This definition is consistent with the eXplainable AI (XAI) literature, where the relevance of an input feature (the object) is generally associated with its impact on the AI model's output (the vehicle's driving) [7]. In addition, it is deliberately framed from a broader driving perspective, so that it can be specialized to more specific safety-related definitions like the ones in [1]-[3] if required by a particular downstream application, while being applicable to a variety of use cases. For example, object relevance could be used to focus or prioritize detection and tracking efforts on objects likely to influence the vehicle's driving, thereby enabling a more efficient allocation of computational resources across the autonomous driving stack. It could also be used to identify the objects that influenced a particular maneuver, making the behavior of an autonomous vehicle easier to inspect and diagnose, enhancing the interpretability of AI-based autonomous driving models.

### *Attention: A Proxy for Relevance?*

The attention mechanism is one of the most popular and widely adopted solutions for estimating the relevance of surrounding objects. Attention is a fundamental component of modern AI architectures. It enables a deep neural network to selectively focus on different parts of the input by learning attention weights that quantify the relative contribution of each input feature to the model's output. Early end-to-end autonomous driving models based on Convolutional Neural Networks (CNNs) and Recurrent Neural Networks (RNNs) leveraged pixel-level attention maps to highlight regions of the input image containing objects that may have influenced the ego-vehicle's driving decisions [8]. An object-level approach to visualize attention in CNN-based models was later proposed in [9], where the authors introduced an additional pooling layer to focus the learned attention weights on individual objects rather than groups of pixels. More recently, pixel-level [10] and object-level attention [3] has also been leveraged to inspect the internal decision-making of autonomous driving models based on more powerful Transformer architectures.

While early studies interpreted attention weights as explanations or proxies for the casual impact of individual input elements on a model's output, studies in the Natural Language Processing (NLP) domain have shown that larger attention weights do not always correlate with greater impact or relevance [11]. In the autonomous driving domain, [8] provided preliminary qualitative evidence challenging the capability of the attention mechanism to identify relevant objects. The authors showed that pixel-level attention maps frequently highlight regions of the input image that do not contain any objects and are, therefore, likely irrelevant to the ego-vehicle's driving. In contrast, [3] and [4] showed that attention can be an accurate proxy for a safety-oriented definition of object relevance considering an autonomous vehicle based on PlanT. In [3], the authors proposed labelling the object receiving the highest attention from PlanT as relevant. However, the accuracy of the proposed approach was indirectly evaluated by analyzing the collision-avoidance capability of a rule-based expert algorithm that, at each timestep, controls the ego-vehicle using only the object labelled as relevant, rather than by directly measuring the relevant object's impact on PlanT's output. In [4], the same approach was evaluated against a human-annotated dataset in which an object is labelled as relevant if the annotator considers it potentially safety-critical to the ego-vehicle's driving. The existing literature provides mixed evidence on the accuracy of attention as a proxy for object relevance. In addition, it highlights the need for a systematic and direct analysis of attention-based relevance estimation methods against the true impact of each individual object on the output of AI-based models such as PlanT, which cannot be reliably determined from rule-based expert algorithms or human annotations. To address these gaps, this work directly analyses the accuracy of multiple attention-based relevance estimation methods against a causal reference of object relevance determined through counterfactual analysis.

### *Counterfactual Analysis: A Causal Reference*

Counterfactual analysis is a fundamental causal learning technique to determine the causal impact of each input feature on a model's output. It involves examining how a model's output changes in response to a controlled modification of one input feature, while keeping all other inputs fixed. This allows the impact of the modified input feature to be isolated from potentially spurious input-output correlations [12], making counterfactual analysis a particularly suitable tool for establishing a causal reference of object relevance against which attention-based methods can be directly evaluated. In autonomous driving, counterfactual analysis has been used, for example, to identify the elements of the driving scene that influence the planned trajectory of an autonomous vehicle [13]. In this work, we perform counterfactual analysis by evaluating an autonomous vehicle's driving under alternative counterfactual scenarios generated by iteratively removing each individual object from the driving scene. For each removed object, we measure the resulting changes in driving behavior and use them as a causal reference for the object's impact on the autonomous vehicle's driving, i.e., its relevance. We perform our analysis in a CARLA-based simulation environment, using an autonomous vehicle controlled by PlanT 2.0 [14], a state-of-the-art, open-source, object-level planning Transformer. CARLA provides the explicit and full control over the driving environment required to remove individual objects while keeping the rest of the driving scene unchanged, enabling the systematic generation of controlled counterfactual scenarios. PlanT 2.0's object-level input representation enables precise counterfactual interventions on individual objects (i.e., object removal), while its attention weights allow the implementation of the attention-based relevance estimation methods considered in this work.

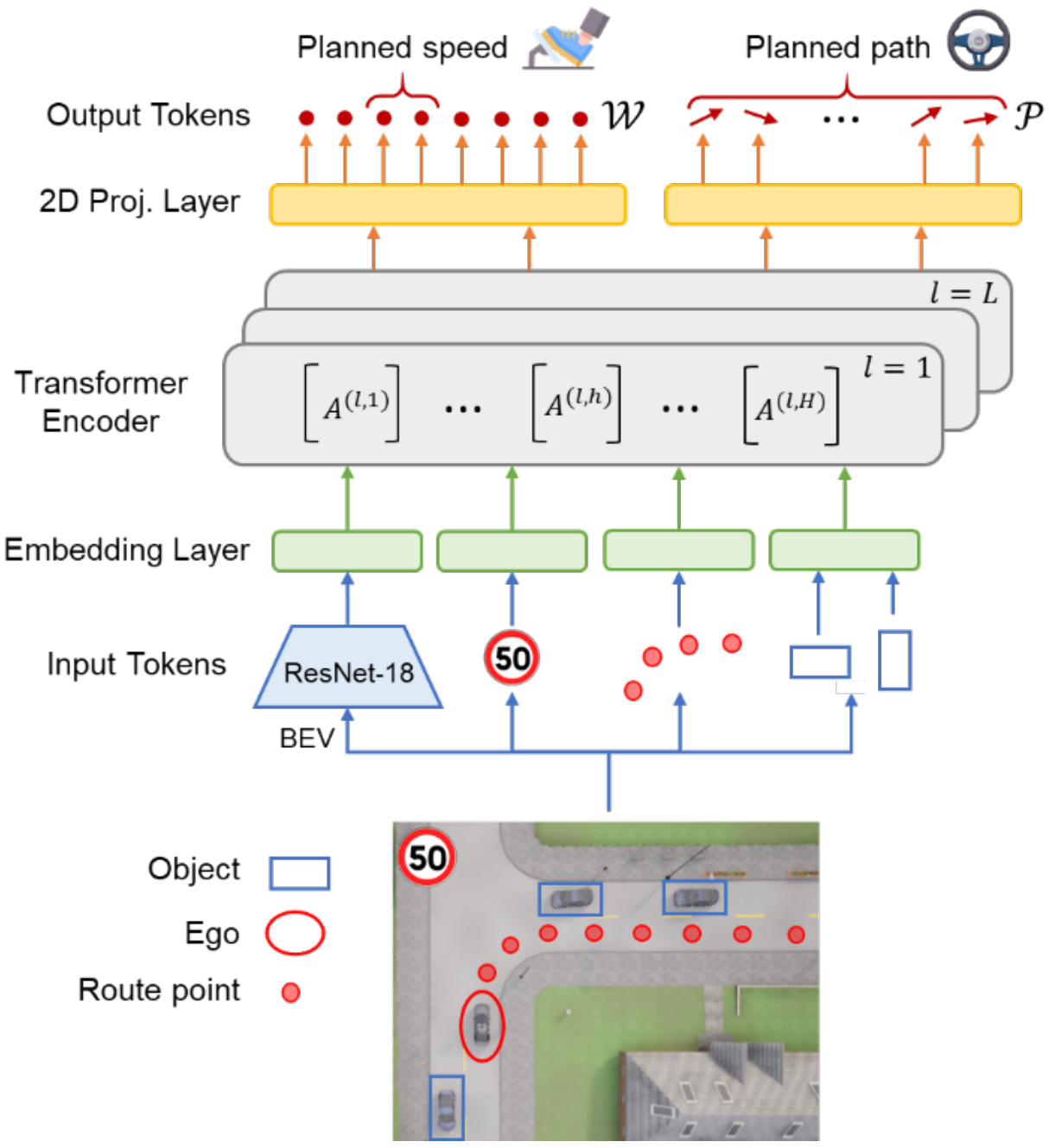

**Figure 1** *PlanT 2.0 architecture.*

## PlanT 2.0

PlanT 2.0 is a Transformer-based model for object-level planning with state-of-the-art performance on popular autonomous driving benchmarks such as Bench2Drive and CARLA's leaderboard 2.0 validation routes at the time of this work. It employs an encoder-only Transformer architecture with 41.4M parameters, $L = 8$ layers, $H = 8$ attention heads per layer, and $d_{model} = 512$ hidden size trained via imitation learning on a dataset of 500k driving scenes collected using the PDM-lite expert driver [14]. PDM-lite is a privileged, rule-based expert planner that uses the kinematic bicycle model to identify potential future collisions and the Intelligent Driver Model (IDM) to compute continuous target velocities based on the distance and relative speed with respect to the object ahead. PDM-lite is widely used to train AI-based autonomous driving models due to its computational efficiency and strong performance on CARLA leaderboard 2.0 scenarios. Further details about PlanT 2.0 training methodology and benchmark performance are available in [14].

Fig. 1 illustrates the architecture of PlanT 2.0. As shown in Fig. 1, PlanT 2.0's input includes multiple types of tokens describing the ego-vehicle's surrounding environment. The first input token represents a Bird Eye View (BEV) map of the road layout centered on the ego-vehicle and is encoded using a ResNet-18 encoder followed by a linear embedding layer. The BEV token is complemented by two learned embedding tokens representing the speed limit and the ego-vehicle's route, respectively. The route is defined by 20 points, sampled every 1 meter, that guide the ego-vehicle's long-term driving behavior towards its goal destination. The remaining input tokens represent surrounding objects. Each object is represented as an oriented 3D bounding box containing its relative position, heading, and class (e.g., vehicle, pedestrian), and is encoded using a class-dependent linear projection layer. PlanT 2.0's output consists of 8 temporally equidistant waypoint tokens sampled at 4 Hz and a sequence of 20 spatially equidistant path tokens sampled every 1 meter. As shown in Fig. 1, both waypoint and path tokens are projected into the 2D driving space using two separate linear projection layers. We denote their respective index sets with $\mathcal{W}$ and $\mathcal{P}$. In PlanT 2.0, the 3rd and 4th waypoint tokens are used to derive the ego-vehicle's planned speed, which controls its longitudinal driving (throttle and braking commands), while all path tokens are used to determine the ego-vehicle's planned path, which controls its lateral driving (steering angle). The attention weights learned by each attention head inside PlanT 2.0's Transformer encoder are stored in a square attention matrix $A^{(l,h)} \in \mathbb{R}^{NxN}$, where $N$ denotes the total number of tokens while $l \in \{1,2,\ldots,L\}$ and $h \in \{1,2,\ldots,H\}$ indicate the $l$-th encoder layer and the $h$-th attention head inside that layer (see Fig. 1). By using $\mathcal{Y}$ to denote the index set of all output tokens (i.e., both waypoint and path tokens), we let $A_{y,o}^{(l,h)}$ denote the $(y,o)$ entry of $A^{(l,h)}$. This entry corresponds to the raw attention weight locally mapping the relative contribution of the $o$-th input object token to the value of the $y$-th output token inside the $h$-th attention head of the $l$-th encoder layer. Note that, in this context, the term contribution refers to how strongly information from the input object is routed to and mixed into the output token's representation through self-attention, rather than to the input object's causal impact on the output token value. We will refer to PlanT 2.0 as PlanT in the rest of this work.

## Methodology

We analyze attention's accuracy as a proxy for object relevance using synthetic driving scenes generated through custom CARLA-SUMO co-simulations. A driving scene consists of a snapshot (or frame) taken from a CARLA simulation including the ego-vehicle, controlled by PlanT, and surrounding objects whose mobility is controlled by SUMO. In each scene, we first identify the relevant objects that influence the ego-vehicle's driving by analyzing their relevance, determined through counterfactual analysis. We then assess the accuracy of attention-based relevance estimation methods by evaluating their ability to correctly classify these objects as relevant. In each driving scene, our analysis considers all PlanT input objects, i.e., all objects within the ego-vehicle's perception range. As in PlanT's original work [14], the perception range is defined by a 50 m-radius circular region behind the ego-vehicle and a 50 m-wide, 100 m-long elliptical region stretching in front of it.

### *Relevance*

We measure the relevance of the $o$-th input object in a driving scene through counterfactual analysis, by quantifying the impact that its removal has on PlanT's planned speed and path, i.e., on the ego-vehicle's longitudinal and lateral driving. Let $v_{ref}$ and $v_o$ denote PlanT's planned speed in the reference driving scene, in which the original set of input objects is not perturbed, and in the counterfactual scenario obtained by removing the $o$-th input object, respectively. We denote the impact of the $o$-th object removal on the planned speed as $\Delta v_o = |\, v_{ref} - v_o|$. Similarly, let $\mathcal{P}_{ref}$ and $\mathcal{P}_o$ denote PlanT's planned path in the reference and counterfactual scenarios. We quantify the impact of the $o$-th object removal on the planned path using the Average Displacement Error (ADE) between $\mathcal{P}_{ref}$ and $\mathcal{P}_o$, denoted by $ADE_o$, which measures the average Euclidean distance between the path points in $\mathcal{P}_{ref}$ and $\mathcal{P}_o$. We define the relevance of the $o$-th input object so that to capture both the longitudinal and lateral impact of its removal on the ego-vehicle's driving:

$$R_o = \max\,(\Delta v_o/\Delta v_{noise},\ ADE_o/ADE_{noise}), \qquad (1)$$

where $\Delta v_{noise}$ and $ADE_{noise}$ represent the noise floors associated with PlanT's sensitivity to the input object order. Since PlanT uses absolute positional embeddings, changing the order of the input objects can cause small, noisy variations in the planned speed or path, even when the set of input objects remains unchanged. Through a series of custom simulations, we numerically characterized the magnitude of these noisy variations and set $\Delta v_{noise}$ = 2.0 m/s and $ADE_{noise}$ = 1.5 m, which correspond to the 99th percentiles of the respective noise-floor distributions.

An input object is relevant if its relevance $R_o$ is greater than 1. According to (1), this means that an input object is relevant if the impact of its removal on the ego vehicle's planned speed or path is larger than the corresponding noise floor. For example, in a single-lane per driving direction scenario, a slower object ahead may cause PlanT to greatly reduce the planned speed of the ego-vehicle without significantly changing its planned path. In this case, the object's impact on the planned speed would lead to $R_o$ > 1, and the object would be considered relevant. In a multi-lane scenario, the same object may instead cause PlanT to change the planned path to let the ego-vehicle change lane and overtake it, potentially with negligible variations to its planned speed. In this case, the object's impact on the planned path would lead to $R_o > 1$ and the object would still be considered relevant.

### *Attention*

Attention-based relevance estimation methods rely on the attention weights learned and stored within PlanT's Transformer encoder to estimate the input objects relevance. They combine attention weights across PlanT's attention heads and layers into an aggregate attention score, $A_{y,o}$, that represents the relative contribution of the $o$-th input object to the $y$-th output token value across the entire model. Online methods compute $A_{y,o}$ using the attention weights obtained with a single forward pass of the model and are therefore suitable for real-time relevance estimation. In contrast, offline methods require additional processing steps, such as gradient backpropagation, which introduce computational overhead and latency, making them more suitable for post-hoc analysis. We focus on the following online methods:

*1) Last:* The aggregate attention score $A_{y,o}$ is equal to the multi-head attention weight in PlanT's last layer, $A_{y,o}^{(L)}$. This method is widely employed in the literature [11], [15]. In general, $A^{(l)}$ denotes the multi-head attention matrix obtained by averaging the attention matrices of all attention heads in PlanT's $l$-th layer, while $A_{y,o}^{(l)}$ denotes its $(y, o)$ entry and $L$ is the index of PlanT's last layer.

*2) Sum:* The aggregate attention score $A_{y,o}$ is obtained by summing the raw attention weights $A_{y,o}^{(l,h)}$ across all attention heads and layers of PlanT. This is the method used by PlanT's authors in their original work [3].

*3) Rollout:* The aggregate attention score $A_{y,o}$ is equal to $\hat{A}_{y,o}$, the $(y, o)$ entry of an attention matrix $\hat{A}$ obtained by recursively multiplying augmented versions of each layer's multi-head attention matrix. Attention rollout [15] is a more sophisticated method that adds the identity matrix to each layer's multi-head attention matrix to account for residual skip connections and uses recursive multiplication to capture the propagation of attention throughout the model.

For each attention-based relevance estimation method, we further combine each aggregate attention score $A_{y,o}$ into a total attention score $A_o$ that captures the contribution of the $o$-th input object across all output tokens influencing the ego-vehicle's driving:

$$A_o = \sum_{y=w\in\{3,4\}} A_{y,o} + \sum_{y=p\in\mathcal{P}} A_{y,o} \,. \qquad (2)$$

The first term corresponds to the aggregate attention score on the 3rd and 4th waypoint tokens, which PlanT uses to determine the ego-vehicle's planned speed and control its longitudinal driving. The second term corresponds to the aggregate attention score on PlanT's planned path tokens, which control the ego-vehicle's lateral driving. Hereafter, we refer to the total attention score $A_o$ of each object simply as the attention score.

In each driving scene, we use the attention scores to identify the objects estimated to be relevant. For each attention-based relevance estimation method, we rank the input objects in decreasing order of their attention score $A_o$, and compute the ratio between each object's attention score and that of the next ranked object. Then, we identify the first ranked object for which this ratio exceeds a pre-defined threshold $\beta$. This object, and all higher-ranked ones, are classified as relevant. Intuitively, this approach assumes that relevant objects will receive higher attention scores than irrelevant ones from PlanT, and that this difference manifests as a drop-off in their attention scores. Unlike existing methods that always classify the highest ranked object as relevant [3], our approach is more flexible: it can identify a single relevant object when the attention score drops after the first ranked object, or multiple relevant objects when the attention score remains high across a group of objects before dropping. In addition, it doesn't necessarily classify an object as relevant when all objects have similar attention scores.

## Attention vs Relevance

We analyze the ability of attention-based relevance estimation methods to identify relevant objects across over 80k synthetic driving scenes generated in CARLA's Town01 and Town10 maps, covering both single- and multi-lane per driving direction scenarios. For each analyzed method, we evaluate recall and precision across different values of $\beta$. In each driving scene, recall is computed as the fraction of relevant objects, identified based on their relevance $R_o$, that are also classified as relevant according to their attention score $A_o$[†]. Recall captures the ability of attention-based methods to identify all relevant objects in a scene and avoid to incorrectly classify a relevant object as irrelevant (false negative). Precision is computed as the fraction of objects classified as relevant based on their attention score $A_o$ that are actually relevant according to their relevance $R_o$. Precision captures the capability of attention-based methods to avoid incorrectly classifying an irrelevant object as relevant (false positive).

Recall and precision are averaged across all driving scenes and reported in Fig. 2(a) and Fig. 2(b), respectively, as a function of $\beta$. Together, Fig. 2(a) and Fig. 2(b) show that attention scores cannot accurately estimate object relevance, challenging the causal interpretation of attention in autonomous driving. This finding is consistent with the broader NLP literature questioning the reliability of attention scores as faithful explanations of a model's output [11]. Specifically, Fig. 2(a) shows that no attention-based relevance estimation method is able to perfectly identify all relevant objects in a scene and achieve a perfect recall, while Fig. 2(b) reveals a strong recall-precision tradeoff across all methods. A lower $\beta$ increases the likelihood of identifying relevant objects (high recall) but also greatly increases the rate of false positives (low precision). Conversely, a higher $\beta$ makes attention-based methods more selective (high precision) but increases the rate of false negatives (low recall). Note that a very small $\beta$ always classifies the highest ranked object in each driving scene as relevant, like in PlanT's original work [3].

[†] Recall is calculated only on driving scenes where there is at least one relevant object.

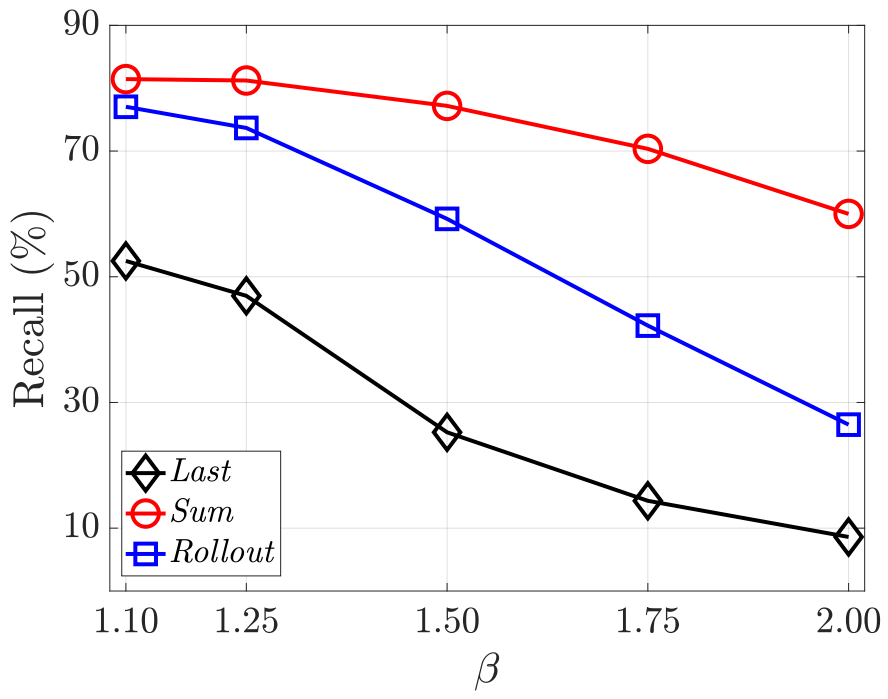

*(a) Recall*

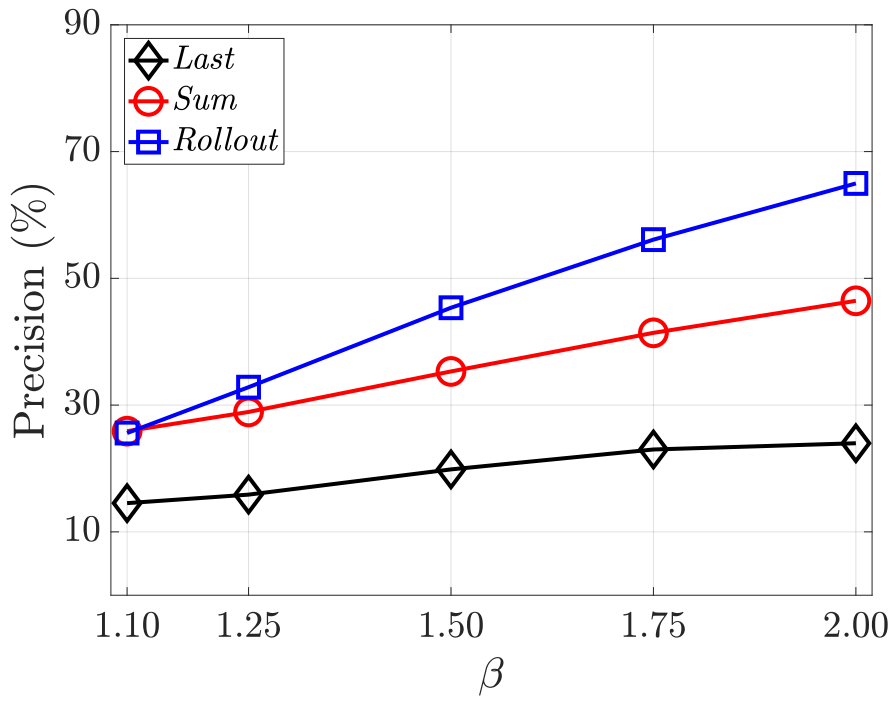

*(b) Precision*

**Figure 2** *Recall (a) and precision (b) of the different attention-based relevance estimation methods, reported as a function of the $\beta$ threshold.*

Fig. 2(a) and Fig. 2(b) show that the *Last* method consistently achieves the lowest recall and precision across all $\beta$ values, indicating that the last-layer attention weights correlate poorly with the objects' relevance measured through counterfactual analysis. Moreover, no method is consistently superior, and *Sum* and *Rollout* exhibit a strong tradeoff between sensitivity (recall) and selectivity (precision). *Rollout*'s multiplicative aggregation tends to assign higher aggregate attention scores only to objects that consistently receive high attention across multiple layers, resulting in a more selective ranking. For this reason, *Rollout* achieves higher precision but may classify as irrelevant those relevant objects with attention weights sparsely distributed across the model (false negative). In contrast, *Sum*'s additive aggregation is less selective and can assign a high aggregate score to an object receiving strong attention only in a few heads or layers. This makes the *Sum* method more capable of identifying relevant objects whose influence on PlanT's output is concentrated in specific parts of the model rather than being consistently distributed throughout it, resulting in higher recall. However, this property also allows irrelevant objects with sparse and uncorrelated attention peaks to accumulate high aggregate scores and be incorrectly classified as relevant (false positive), thereby lowering precision.

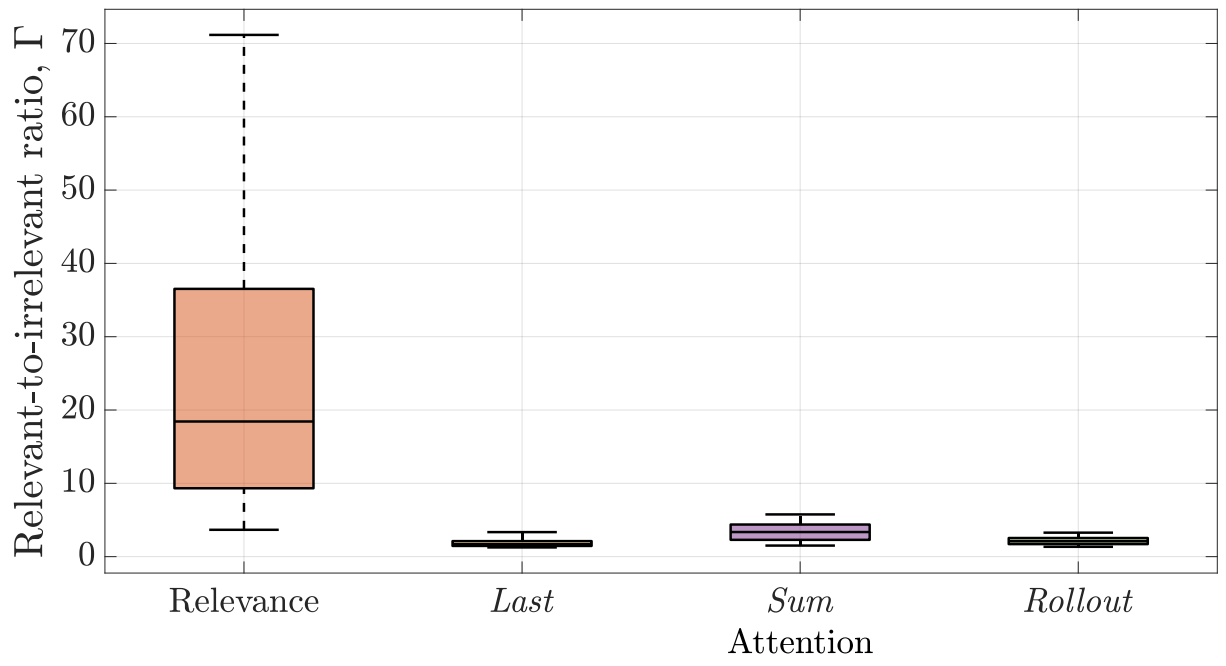

**Figure 3** *Relevant-to-irrelevant ratio, $\Gamma$, for $\beta$ = 1.1. Boxplots' whiskers indicate the 5th and 95th percentile.*

We further compare the three attention-based relevance estimation methods in terms of their ability to discriminate between relevant and irrelevant objects. To this end, we introduce the relevant-to-irrelevant ratio $\Gamma$. For each relevant object, we define $\Gamma$ as the ratio between its relevance $R_o$ and the average relevance of all irrelevant objects in the same scene. For each object classified as relevant by attention-based methods, $\Gamma$ is analogously defined as the ratio between its attention score $A_o$ and the average attention score of all objects classified as irrelevant. We compute $\Gamma$ only on driving scenes where all three attention-based methods correctly identify all relevant objects to ensure a fair comparison. Fig. 3 shows the distribution of the $\Gamma$ values measured for $\beta = 1.1$, which corresponds to the highest-recall operating point for all three attention-based methods in Fig. 2(a). The results show that, even when attention correctly identifies all relevant objects, it provides a substantially weaker discrimination between relevant and irrelevant objects than relevance. At the 95th percentile, relevance is characterized by $\Gamma$ values that are approximately ten times larger than those of attention-based methods. This gap indicates that PlanT's attention weights are not strongly concentrated on relevant objects, but tend to be distributed across all input objects, thus limiting the separation between relevant and irrelevant objects. In contrast, relevance assigns negligible values to irrelevant objects by design and provides a more discriminative and interpretable measure of the impact that each input object has on PlanT's output and, ultimately, the ego-vehicle's driving.

### *Analysis by Driving Scenario*

We further investigate the ability of attention-based relevance estimation methods to identify relevant objects by analyzing their recall and precision in straight and intersection scenarios, considering both single and multiple lane per driving direction road topologies. An intersection scenario is one in which the ego-vehicle is approaching an intersection and is within 50 m from its center; all other scenarios are classified as straight. Table 1 reports the recall and precision achieved by the *Sum* method across all scenarios when $\beta = 1.1$ and $\beta = 2.0$, respectively. These thresholds correspond to *Sum*'s highest recall and precision values in Fig. 2. For each scene, we denote by $N$ and $M$ the number of relevant objects and the number of objects classified as relevant by the *Sum* method, respectively.

| Scenario | Recall ($\beta = 1.1$) | | Precision ($\beta = 2.0$) | |
|---|---|---|---|---|
| | $N = 1$ | $N = 2$ | $M = 1$ | $M = 2$ |
| Single-lane, straight | 98.3% | - | 27.9% | 25.8% |
| Multi-lane, straight | 90.5% | 48.6% | 25.1% | 18.5% |
| Single-lane, intersection | 85.5% | 48.1% | 52.0% | 49.4% |
| Multi-lane, intersection | 85.3% | 47.6% | 50.9% | 47.3% |

**Table 1** *Highest recall and precision for the Sum method in single- and multi-lane scenarios.*

Table 1 shows that recall is very high (98.3%) in straight single-lane scenarios with one relevant object ($N = 1$), indicating that attention can consistently identify relevant objects in simple scenarios. Recall for $N = 2$ is not reported in Table 1 because no single-lane straight scenario in the dataset contains more than one relevant object. This result is consistent with the intuitive expectation that, in such simple scenarios, the ego-vehicle's driving is influenced only by the object directly ahead. The high recall observed for $N = 1$ mainly stems from attention's tendency to classify the object in front of the ego-vehicle as relevant. This tendency may be related to PlanT's training methodology: PlanT is trained to imitate an expert driver that relies on the IDM and, therefore, adjusts its speed focusing on the object directly ahead. However, strongly focusing on the object ahead can reduce precision, since object relevance depends not only on the relative position between the ego-vehicle and the object but also on other factors, such as their relative speed. Indeed, precision is only 27.9% in single-lane straight scenarios when $M = 1$.

This tendency to focus on the object ahead becomes less effective in more complex scenarios where the ego-vehicle's driving can be influenced by richer interactions with multiple surrounding objects. In multi-lane straight scenarios, recall decreases to 90.5% for $N = 1$ and further drops to 48.6% for $N = 2$, highlighting the limited ability of attention-based methods to identify multiple relevant objects in the same scene. Fig. 4(a) illustrates this limitation with an example in which attention identifies only one of the two relevant objects in front of the ego-vehicle. In the figure, the ego-vehicle is highlighted by a red circle, relevant objects by light blue dashed boxes, and the object classified as relevant by the *Sum* method by a green box. Arrows indicate the direction of the ego-vehicle's planned path and the objects' trajectories, which are not used by PlanT as input. In Fig. 4(a), both objects ahead of the ego-vehicle are relevant since they are moving at lower speed and blocking both lanes, forcing the ego-vehicle to decelerate. However, attention correctly classifies as relevant only the object in the ego-vehicle's current lane.

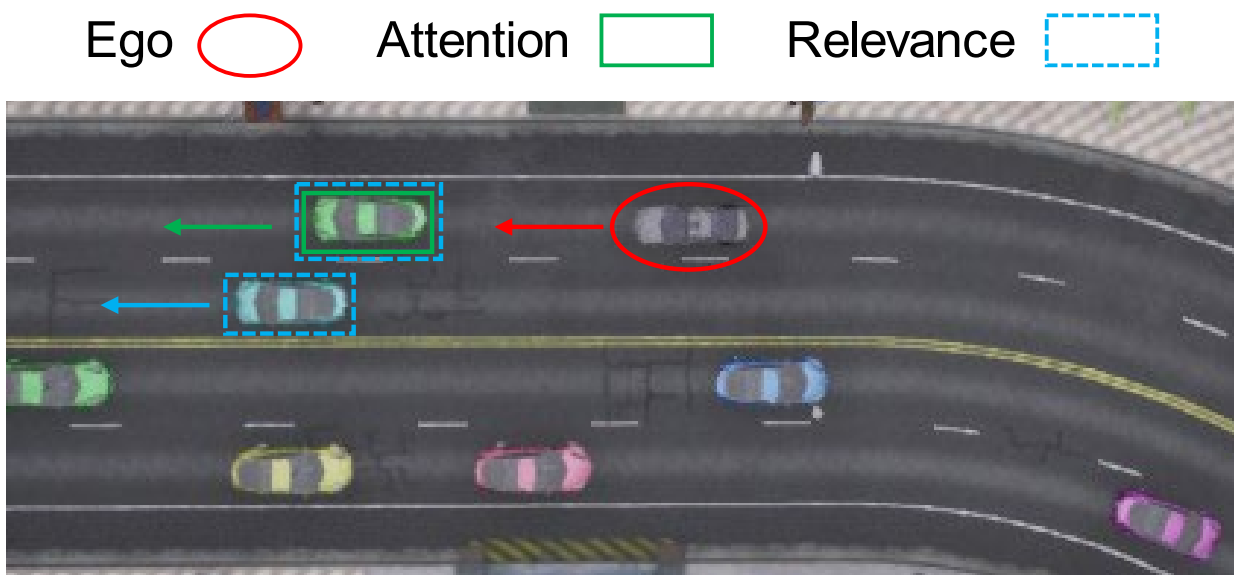

*(a) Multi-lane intersection scenario with bad recall: attention identifies only one of the two relevant objects.*

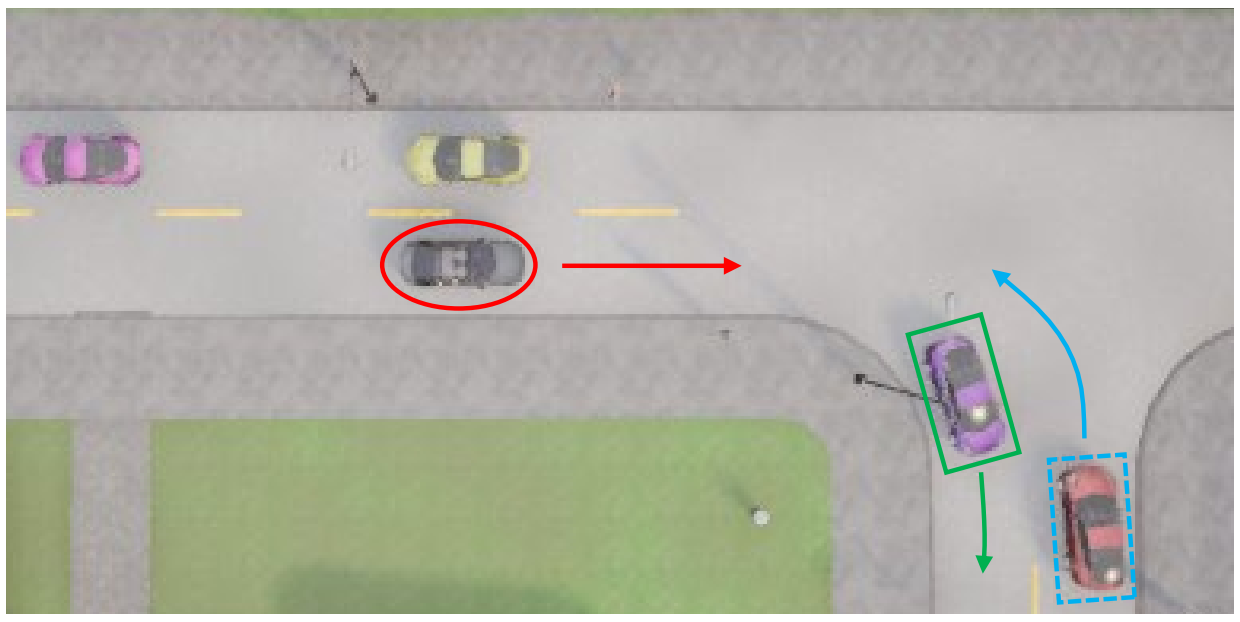

*(b) Intersection scenario with bad precision and recall: attention focuses on the object in front (green box) and misses the relevant one (light blue box).*

**Figure 4** *Qualitative examples for single-lane and multi-lane scenarios.*

Attention's limited ability to identify relevant objects in complex settings is further highlighted by the analysis of intersection scenarios. Fig. 4(b) shows a representative example in which attention focuses on the irrelevant object directly ahead of the ego-vehicle (in the green box), which is leaving the intersection, while failing to classify as relevant the potentially safety-critical object (in the light blue box) crossing the ego-vehicle's path. In both single- and multi-lane intersection scenarios, recall drops to approximately 85% when $N = 1$, and decreases further in scenarios with two relevant objects ($N = 2$).

The similar performance in single- and multi-lane intersection scenarios can be explained by the fact that they are approximately equivalent from the ego-vehicle's driving perspective, since the ego-vehicle is allowed to cross multi-lane intersections only on the innermost lane. Table 1 also shows that precision is generally higher in intersections compared to straight scenarios. This can be explained by the richer interactions and greater number of potential conflicts at intersections, which increases the likelihood that an object classified as relevant is actually affecting PlanT's output and, hence, the ego-vehicle's driving.

## Conclusions

This work analyzed attention's accuracy in estimating the relevance of surrounding objects in autonomous driving. To this end, we analyzed more than 80k synthetic driving scenes with an autonomous vehicle based on PlanT 2.0, a state-of-the-art, object-level planning Transformer. Our analysis shows that attention is a correlational rather than causal measure of object relevance, corroborating preliminary evidence from the NLP domain. In particular, the obtained results show that attention tends to rely on learned correlations stemming from the autonomous vehicle's training methodology, rather than accurately capturing each object's causal impact on the vehicle's driving. While attention-based relevance estimation methods can identify relevant objects in simple straight-road scenarios, they do not generalize to more complex scenarios with multiple lanes per driving direction and richer interactions, such as intersections. Furthermore, they tend to incorrectly classify as relevant objects with little or no impact on the vehicle's driving. Results also show that attention-based relevance estimation methods are not able to strongly discriminate between relevant and irrelevant objects, and exhibit an inherent tradeoff between sensitivity (recall) and selectivity (precision). Our findings highlight the limitations of attention as a proxy for object relevance in autonomous driving, underscoring the need for more accurate proxies able to capture the causal impact of individual objects on the vehicle's driving and distinguish genuine decision-making influence from correlational associations.

*Luca Lusvarghi is a senior researcher at the Networked Systems Laboratory of the Universidad Miguel Hernández (UMH), Elche, Spain. His main research interests are centred on semantic and task-oriented V2X communications as well as connected and autonomous driving. He was the recipient of the GTTI award for the best Italian PhD thesis in ICT and of a 2023 MSCA postdoctoral fellowship (SemanticV2X project).*

*Javier Gozalvez is a full professor and director of the Networked Systems Laboratory at the Universidad Miguel Hernández (UMH), Elche, Spain, where he leads research activities in the areas of vehicular networks, connected and autonomous driving, 6G, and industrial networks. He is an IEEE Fellow, served as President of the IEEE Vehicular Technology Society, Editor in Chief of the IEEE Vehicular Technology Magazine, Distinguished Lecturer and Distinguished Speaker for IEEE VTS.*